\documentclass[letterpaper]{article} 
\usepackage{aaai25}  
\usepackage{times}  
\usepackage{helvet}  
\usepackage{courier}  
\usepackage[hyphens]{url}  
\usepackage{graphicx} 
\usepackage{natbib}  
\usepackage{caption} 
\usepackage{algorithm}
\usepackage{algorithmic}

\usepackage{xcolor}
\newcommand{\answerYes}[1]{\textcolor{blue}{#1}} 
\newcommand{\answerNo}[1]{\textcolor{teal}{#1}} 
\newcommand{\answerNA}[1]{\textcolor{gray}{#1}}

\usepackage{newfloat}
\usepackage{listings}
\DeclareCaptionStyle{ruled}{labelfont=normalfont,labelsep=colon,strut=off} 
\floatstyle{ruled}
\newfloat{listing}{tb}{lst}{}
\floatname{listing}{Listing}
\title{The Emerging Generative Artificial Intelligence Divide in the United States}
\author {
    Madeleine I. G. Daepp\textsuperscript{\rm 1},
    Scott Counts\textsuperscript{\rm 1}
}
\affiliations {
    \textsuperscript{\rm 1}Microsoft Research\\
    mdaepp@microsoft.com, counts@microsoft.com
}

\begin{document}

\maketitle

\begin{abstract}
\textcolor{red}{This is the author‑accepted version. Please cite the AAAI ICWSM’25 proceedings version when available.}
The digital divide refers to disparities in access to and use of digital tooling across social and economic groups. This divide can reinforce marginalization both at the individual level and at the level of places, because persistent economic advantages accrue to places where new technologies are adopted early.  To what extent are emerging generative artificial intelligence (AI) tools subject to these social and spatial divides? We leverage a large-scale search query database to characterize U.S. residents' knowledge of a novel generative AI tool, ChatGPT, during its first six months of release. We identify hotspots of higher-than-expected search volumes for ChatGPT in coastal metropolitan areas, while coldspots are evident in the American South, Appalachia, and the Midwest. Nationwide, counties with the highest rates of search have proportionally more educated and more economically advantaged populations, as well as proportionally more technology and finance-sector jobs in comparison with other counties or with the national average. Observed associations with race/ethnicity and urbanicity are attenuated in fully adjusted hierarchical models, but education emerges as the strongest positive predictor of generative AI awareness. In the absence of intervention, early differences in uptake show a potential to reinforce existing spatial and socioeconomic divides.
\end{abstract}

\section{Introduction}

Generative artificial intelligence (AI) is a general-purpose technology, an innovation that is affecting many sectors of the economy~\citep{eloundou2023gpts}. Previous general-purpose technologies took decades to centuries to achieve widespread adoption~\citep{bresnahan1995general, lipsey2005economic}. By contrast, several generative AI models have now been made freely available, within months of development, to people across the globe, so long as they have access to a device with a stable internet connection~\citep{gordon2022}. Hypothetically, then, a plurality of people should be accessing these new tools. But adoption could also be subject to the ``digital divide'', differences between those who do and do not benefit from digital tooling that can reinforce social and economic marginalization in knowledge economies~\citep{vandijk2005deepening, hargittai2018digital}. Because persistent economic advantages accrue to places where new technologies are adopted early~\citep{saxenian1996regional, autor2013growth, moretti2012new, moretti2022place}, it is important to understand how adoption differs across places as well as across groups.

There is extensive prior literature documenting digital divides in knowledge and use of technologies including personal computers~\citep{chakraborty2005measuring}, internet~\citep{scheerder2019negative}, task-specific AI~\citep{lutz2019digital}, and algorithmic AI~\citep{oeldorf2023we, wang2024artificial}. However, little research has characterized the digital divide with respect to generative AI, models that can create new content without explicit programming beyond initial guidelines (prompts)~\citep{solaiman2023evaluating, khowaja2024chatgpt}. 

Early differences in access to and use of new generative AI tooling could have important and long-lasting social implications. Experimental studies show that access to generative AI can have large effects on productivity~\citep{noy2023experimental, brynjolfsson2023generative, peng2023impact}, creativity~\citep{zhou2023generative, dell2023navigating, girotra2023ideas, doshi2023generative}, and performance~\citep{choi2023ai, brynjolfsson2023generative, dell2023navigating}, with several studies finding the largest benefits for the least-skilled users~\citep{noy2023experimental, brynjolfsson2023generative, dell2023navigating}. Disparities in adoption could preclude these tools from actually reaching those who would most benefit: several individual-level surveys find that generative AI's adoption is higher among younger, more educated, and higher income people and lags in other populations~\citep{bick2024rapid, Motyl2024, ParkGellesWatnick2023}; at the level of countries, researchers have observed strong positive associations between awareness of generative AI and human capital~\citep{pahl2024} or income~\citep{khowaja2024chatgpt, liu2024earth}. These patterns raise concerns about emerging social and geographic divides in access---but they may also reflect genuine variation in perceived usefulness across places with different industry and occupational structures. There is thus a need for detailed within-country analysis to better characterize these dynamics.

This study helps to address this research gap, using a large-scale search log database to analyze awareness of generative AI across the United States. We take as a case study the first six months after the release of ChatGPT, a simple chat interface atop a powerful large language model, which saw rapid uptake after its initial public release on November 30, 2022~\citep{chatgpt2023100mill}. We ask three core questions:
\begin{enumerate}
    \item Where is awareness of generative AI, as proxied by search interest in ChatGPT, most versus least concentrated?
    \item What demographic and socioeconomic factors are associated with higher versus lower levels of awareness?
    \item To what extent are associations with socioeconomic factors accounted for by differences in sectoral makeup?
\end{enumerate}

To answer these questions, we characterize the fraction of internet searches for ChatGPT by county, documenting statistically significant clustering in the counties with the highest versus lowest rates of search for ChatGPT. Counties with the highest rates of search are proportionally more urbanized, more educated, and have more technology and creative sector jobs in comparison with counties with lower rates of search or the U.S. overall.  In fully adjusted hierarchical models, associations with demographic factors and sectoral makeup are attenuated after accounting for educational attainment---a finding that is consistent with prior literature on digital divides~\citep{blank2016dimensions, goel2012does, van2014digital, elena2021assessing}. 

Through this work, we make three contributions: 
\begin{itemize}
\item First, we use a large-scale search log database to characterize emerging disparities with respect to awareness of a major new digital tool in the first six months of its public release.
\item Second, we apply spatial clustering methods to identify hotspots as well as to surface coldspots that could be prioritized for intervention.
\item Finally, we highlight educational attainment as a key differentiating factor between places with high versus lower search rates. This finding offers an early warning that disparities in access are following the patterns established in prior divides.
\end{itemize}

\section{Related Research}

The digital divide refers to social and economic disparities with respect to technology at multiple levels~\citep{wei2011conceptualizing, hargittai2002second, van2006digital, van2020digital, robinson2020digital}. The first, foundational level of the digital divide is the question of access: whether a person has an internet connection or the devices required to access it~\citep{oecd2001}. Divides in access track preexisting social and economic disparities with respect to factors like income, education, race, age, and gender in the U.S. and globally~\citep{dimaggio2004digital, van2006digital, mubarak2020confirming}. Mobile phones and reduced data costs have helped to bring over two-thirds of the world's population online~\citep{ITU2023}, however, and over 90\% of U.S. adults are now estimated to have internet access~\citep{Gelles-Watnick2024}. 

Simply having access, however, is far from adequate to ensure effective or beneficial use. \citet{hargittai2002second} describes a second-level divide in skills and usage that remains evident even in highly digitized societies~\citep{dimaggio2004digital, van2006digital, hargittai2018digital}. Researchers document significant disparities in skills, usage, and attitudes toward digital technologies between social or economic groups, with studies consistently showing that the least skilled groups are older and less educated compared to other groups~\citep{vanDeursen2019, elena2021assessing, scheerder2017determinants}. \citet{scheerder2017determinants} further emphasize the importance of a third-level divide in which groups differentially experience benefits versus harms from their use of digital tooling. Families with less versus more education are more likely, for example, to struggle to cope with fraud or scams online~\citep{scheerder2019negative} and people from minoritized racial groups are more likely than majority groups to be inadequately served or actively surveilled by emerging technologies~\citep{zuboff2019age, benjamin2019race, noble2018algorithms}.

Despite this rich literature on prior digital divides, research on disparities in generative AI's adoption remains limited. Scholars have documented significant disparities in AI knowledge, attitudes, and impacts~\citep{wang2024artificial, ragnedda2020new, cotter2020algorithmic, gran2021or, zarouali2021investigating, park2022digital}, but these studies have tended to focus on AI as implemented via algorithmic recommender systems~\citep{oeldorf2023we} and interactive devices~\citep{lutz2019digital}. 
It is unclear whether the patterns they surface will be similar for the case of generative AI---AI that creates new content in response to user prompting~\citep{solaiman2023evaluating}---given increasing evidence of generative AI's potential to reduce skill and productivity gaps~\citep{noy2023experimental, brynjolfsson2023generative, dell2023navigating}. Racial or ethnic digital divides due to differences in needs, norms, and power are likely, in the absence of intervention, to be self-reinforcing due to their encoding in training data and model development~\citep{raji2020closing, noble2018algorithms}. Moreover, while the existing digital divide literature tends to emphasize social and economic divides, there is also a need to examine differences in adoption across places because, as economic development research shows, early differences in the adoption of novel technologies across places can foster persistent differences in residents' wealth and well-being over time~\citep{saxenian1996regional, autor2013growth, moretti2012new, moretti2022place}. 

Emerging evidence suggests strong associations between adoption and economic advantage. Examining global divides, \citet{khowaja2024chatgpt} document that nearly two thirds of traffic to the ChatGPT website is sourced from high-income countries. Within the U.S., \citet{bick2024rapid} document positive individual-level associations between adoption rates and income or education levels, a finding that is consistent with the correlations observed in other population-representative surveys~\citep{Motyl2024, ParkGellesWatnick2023}. While these surveys find that less than 20\% of US adults reported using AI-based text generation tooling in 2023, a study of professionals estimated that more than 90\% of US-based computer programmers in enterprise companies were using generative AI~\citep{Shani2023}---offering a warning sign of uneven adoption. To our knowledge, however, there have not yet been systematic investigations of the spatial distribution of these associations, or the extent to which associations persist after accounting for confounding factors. This study seeks to fill the research gap, documenting emerging evidence of significant spatial and socioeconomic disparities in awareness of a major AI tool that are similar to the disparities that have characterized past digital divides.

\section{Methods}

\subsection{Data Collection}

Our dataset comprises de-identified interactions from Microsoft's Bing search engine. We analyze billions of searches collected between December 1, 2022 and May 31, 2023 in the U.S., including both desktop and mobile query search. All data were de-identified, aggregated to the ZIP code and then to the county level, and securely stored. Data were analyzed only after aggregation to preserve user privacy and in accordance with Bing's privacy policy. The study was reviewed and approved by the Microsoft Research Institutional Review Board (Protocol ID 10590) prior to research activities.

To construct our dataset, we first calculated the counts of searches for ChatGPT as well as total search counts and user counts by county. Each interaction includes the search query string as well as a randomly generated client ID, a timestamp, and the associated ZIP code and state. We added an indicator of whether the query referred to ChatGPT by a case-insensitive search for the terms ``chatgpt'' or the similarly common ``chat gpt''. We then aggregated total counts and unique user counts by ZIP code. We limited our keyword set intentionally to maintain high precision, avoiding potentially related terms like ``chat'' and ``AI'' that, upon initial inspection, included a high false positive rate. The ZIP code linkage is provided via a proprietary location inference engine that uses contextual and historical information to improve upon accuracy over standard reverse IP lookup databases. We linked ZIP codes with counties using the HUD-USPS ZIP Code crosswalk files~\citep{din2020crosswalking}. For any ZIP code linked with multiple counties, we allocated queries to counties proportionally to the ratio of all addresses in the ZIP-county overlap to the total number of all addresses in the ZIP code. Finally, we suppressed data from county-query cells with fewer than 50 unique users to preserve anonymity. Our final data set comprises 2,397 counties.\footnote{After suppression, our data set captured 76\% of all U.S. counties, covering 98\% of the total U.S. population (because the suppressed counties were sparsely populated). We also aggregated counts of overall and ChatGPT queries by state and month; at these levels of aggregation, all cells meet the threshold for inclusion.} 

We used Bing data because of its broad national coverage. Using proprietary search engine data enabled us to examine finer spatial geographies than those available using comparable, publicly available datasets, following~\citet{suh2022disparate}. As a robustness check, we compared state-level rates of search for ChatGPT with publicly available Google Trends query data~\citep{google_trends_2023}. The two datasets are highly correlated (Pearson's Correlation = 0.86). 

Finally, we obtained county-level socioeconomic, demographic, and industry makeup from the 5-year American Community Survey for 2016-2020 using the National Historical Geographic Information System~\citep{manson2023ipums}. We used these data to construct socioeconomic variables including: percent college educated, calculated as the fraction of the population over 25 with a college education or higher; percent rural, or the fraction of the population living in a rural area; median household income in the past 12 months; and the unemployment rate, or the fraction of the civilian labor force 16 years and over that is unemployed. To characterize demographic makeup, we first calculated Hispanic or Latino as a fraction of the total population; for the other three common demographic groups (White, Black, and Asian) we calculated the percent identifying with that group only as a fraction of the total population. We additionally calculated industry makeup as the percent of all people in the adult employed civilian population who are employed in technology, arts, finance, or service sector jobs. 

\subsection{Spatial Analysis}

We first mapped logged rates of search for ChatGPT by state. At this level of aggregation, we were also able to evaluate trends over time, examining rates over each month of observation. We further assessed univariate associations of search rates and county-level socioeconomic, demographic, and industry makeup. Because county-level correlates have skewed and non-normal distributions, we examined logged rates of search for ChatGPT in comparison with counties' percentile ranks for each variable, calculating medians and interquartile ranges (IQRs) of search rates across counties, weighted by population, in each percentile rank bin. 

We evaluated spatial clustering of county-level search rates using the Moran's I statistic, which tests the extent to which there is more spatial clustering of observed values relative to what would be expected if values were randomly distributed~\citep{moran1950notes}. The Moran's I statistic is normalized so that values fall between -1 and 1, with positive values indicating that neighboring features tend to be both smaller or both larger than the mean (clustering) and negative values indicating that features higher than the mean tend to be located next to features with low scores (dispersal).

After confirming the presence of global spatial autocorrelation, we mapped local clustering patterns by calculating the Getis-Ord G* statistic~\citep{getis1992analysis}. The G* statistic can be used to test the null hypothesis that neighboring counties have rates no more similar to one another than would be expected by random chance. In contrast with the Moran's I, which is a global score across all counties, the G* statistic can be calculated for each county individually~\citep{anselin1995local}. We thus mapped results using a two-sided 5\% cutoff, identifying ``hotspots'' as counties where the G* statistic $>$ 1.96 and ``coldspots'' as those counties where the G* statistic $<$ -1.96. 

\subsection{Hierarchical Modeling}

We used multivariate negative binomial regression to estimate the association of search rates with socioeconomic and demographic factors net of the effects of industry makeup. To account for the spatial autocorrelation in our data set, we used a multilevel specification, allowing a random effect by state.

We fit models with the sets of socioeconomic, demographic, and industry makeup variables separately; we then fit a full specification including all covariates. We standardized and scaled covariates to stabilize the model fit and to facilitate comparison across coefficients. After observing large changes in the full model in comparison with the separate specifications---consistent with confounding---we iteratively removed regressors to evaluate the change across other covariates associated with each variable's exclusion. Finally, we assessed the fit of the final model by mapping the predicted and residual values and again using the Moran's I statistic to test for residual spatial autocorrelation.\footnote{We conducted our analyses in R 4.1.1~\citep{R_4.1.1}, using the sf~\citep{pebesma2018}, usmap~\citep{dilorenzo2023usmap}, hmisc~\citep{hmisc}, and spdep~\citep{bivand} packages for spatial analysis and the glmmTMB~\citep{glmmtmb} and DHARMa packages~\citep{DHARMa} to fit and assess models.}

\section{Results}

\begin{figure}[t]
\centering
\includegraphics[width=0.95\columnwidth]{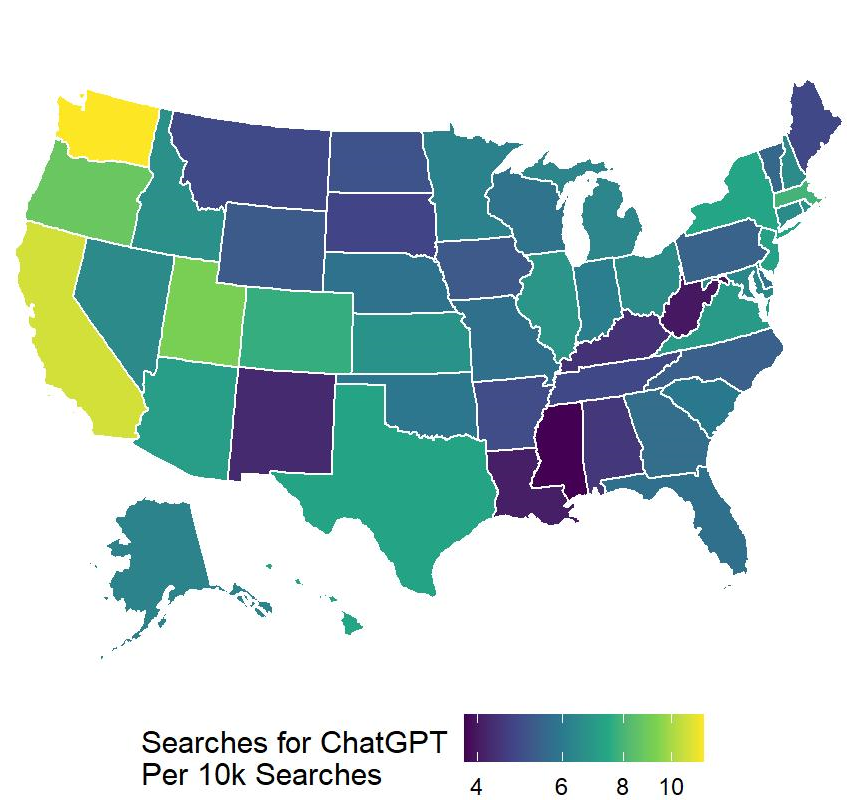}
\caption{Rates of Search for ChatGPT by State. Colors indicate the number of ChatGPT references per 10k searches in the first six months since the tool's initial public release.}\label{fig1:rawmaps}
\end{figure}

Over the six months of observation, we calculate a median rate of approximately 6.2 searches for ChatGPT per 10,000 total searches (IQR 4.5 - 7.7).  Figure~\ref{fig1:rawmaps} shows rates of search for ChatGPT by state. Search rates are highest in Washington and California and lowest in states in the Gulf (Louisiana, Alabama, Mississippi) and Appalachia (West Virginia, Kentucky and Tennessee), as well as New Mexico and South Dakota.

\begin{figure}[htbp]%
\centering
\includegraphics[width=0.95\columnwidth]{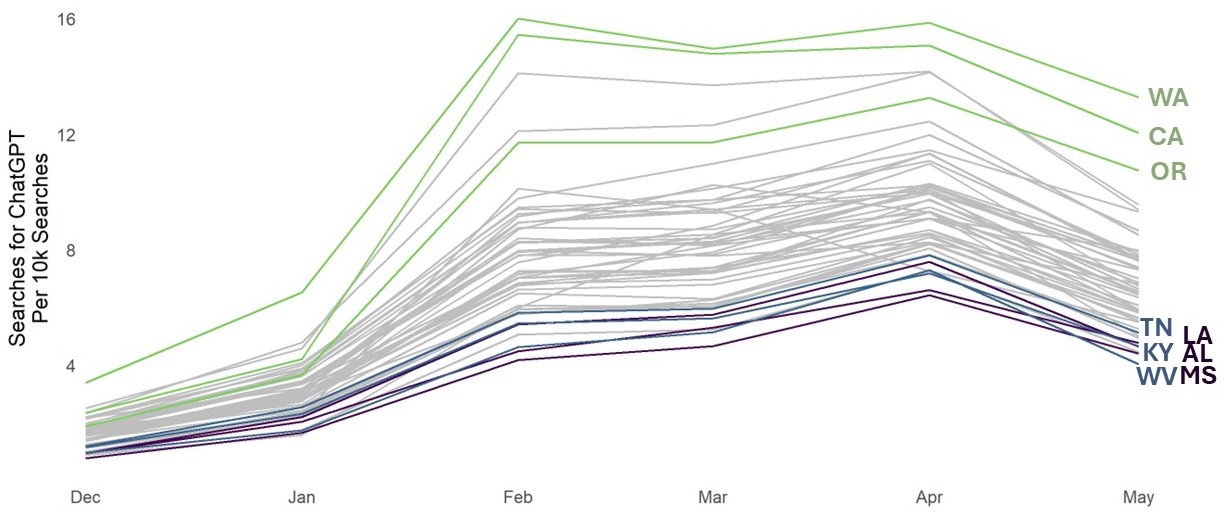}
\caption{Average monthly search rates for ChatGPT (per 10k searches). West Coast states with the highest rates---Washington (WA), California (CA), Oregon (OR)---appear in green. States with persistently low search rates are highlighted in purple for the Gulf South---Louisiana (LA), Alabama (AL), Mississippi (MS)---and blue for Appalachia---Tennessee (TN), West Virginia (WV), Kentucky (KY).}\label{fig2:timeline}
\end{figure}

These spatial differences emerge early and persist throughout the observation period. Figure~\ref{fig2:timeline} shows state-level search rates disaggregated by month. The West Coast states rank in the top 10\% of all state-period observations beginning in February, and remain among the highest-ranked regions through May.\footnote{The remaining two places in the top five with respect to search rates are the District of Columbia and Utah.} Although rates of search for ChatGPT do increase in Gulf and Appalachian states over this period, these states persistently rank in the bottom 10 states with respect to interest over each month of the six-month observation period.

\begin{figure}[htbp]%
\centering
\includegraphics[width=0.95\columnwidth]{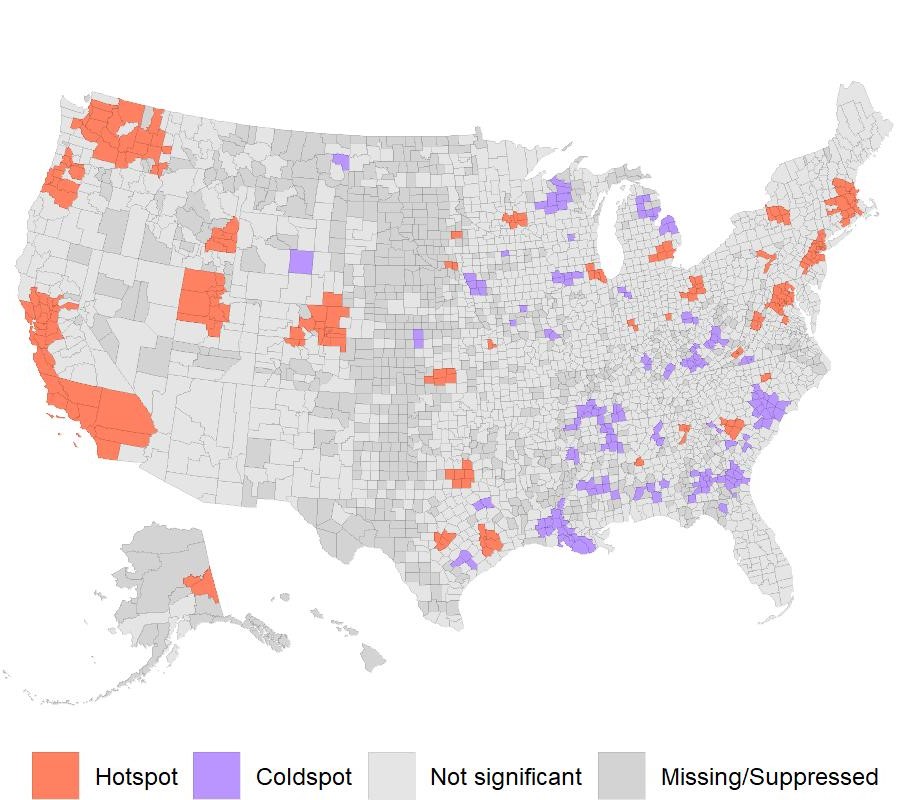}
\caption{Detection of hot- and coldspots with the Getis-Ord G* Statistic. Red indicates the presence of a statistically significant hotspot (G* $>$ 1.96) and blue indicates the presence of a statistically significant coldspot (G* $<$ -1.96).}\label{fig3:clusters}
\end{figure}

\begin{figure*}[t]
\centering
\includegraphics[width=0.9\textwidth]{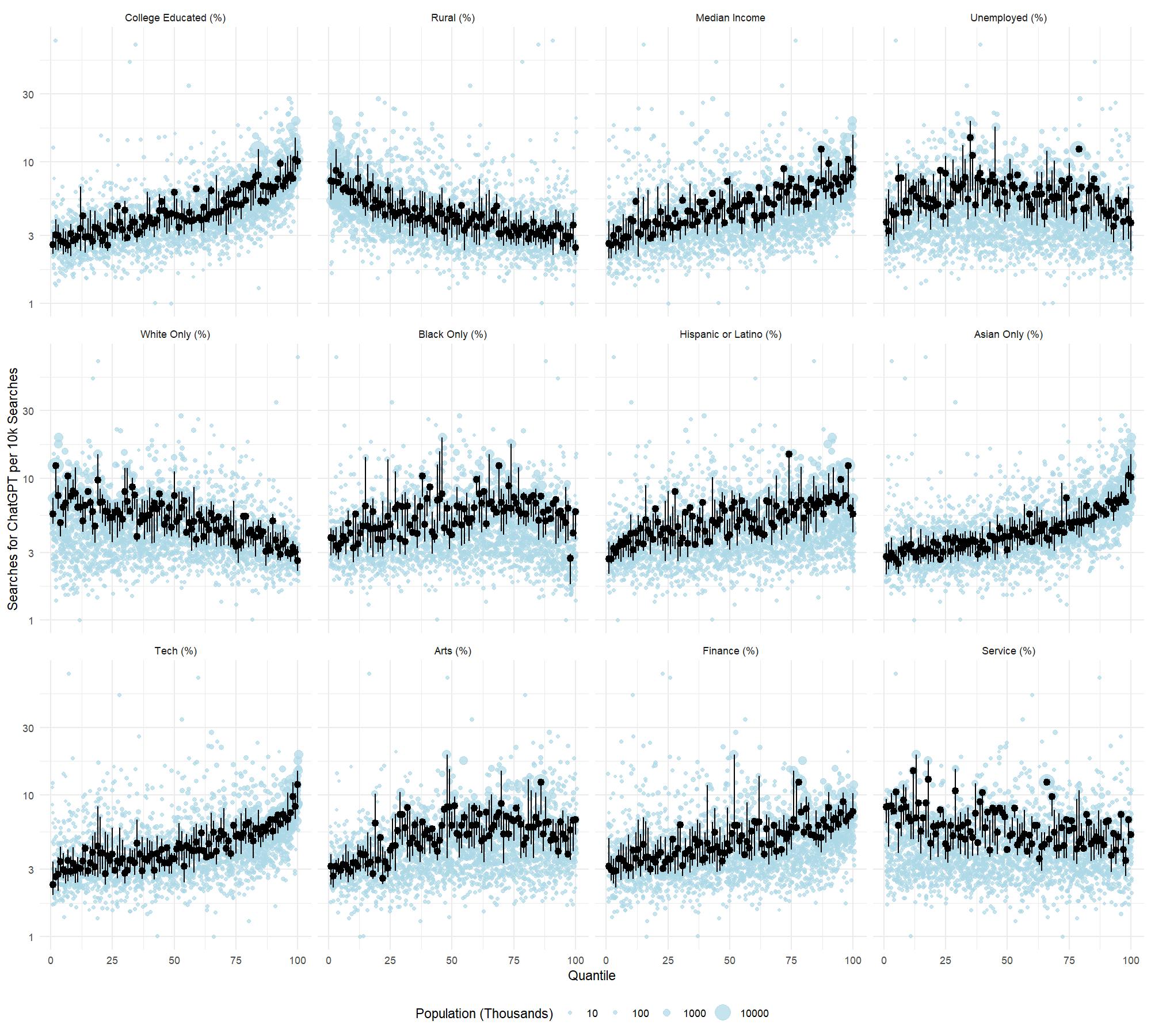}
\caption{Searches for ChatGPT in relation to socioeconomic, demographic, and sector makeup. Blue points indicate raw observations; black points are medians and 75\% interquartile ranges. Estimates are weighted by county population.}\label{fig4:all}
\end{figure*}

Examining rates of search at the county level, we observe a moderate but statistically significant level of spatial clustering (Moran's I = 0.26, p $<$ 0.001). This result suggests that counties with higher rates of search for ChatGPT are likely to be geographically clustered. The estimate is robust to the exclusion of values from outlier counties: examining winsorized rather than raw values, we see a similar result (Moran's I = 0.27, p $<$ 0.001). 

Figure~\ref{fig3:clusters} shows clusters of neighboring counties with more elevated rates of search (hotspots) or lower rates of search (coldspots) than would be expected if rates were distributed randomly; we note that the hotspots coincide with major metropolitan areas in the West (Seattle, Portland, San Francisco and Los Angeles); in the Mountain West (Salt Lake City and Denver); on the East coast (Boston, New York City, Philadelphia, and Washington D.C); and in Texas (Houston, Austin, and Dallas) and Atlanta. Coldspots are predominantly observed in the Gulf South, Appalachia, and the rural Midwest, though several Midwestern metropolitan areas do emerge as hotspots (Chicago, Minneapolis, and Detroit). 

\begin{table*}[t]
\centering
\begin{tabular}[t]{lllllllll}
\hline
 & Model 1 & Model 2 & Model 3 & Model 4 & Model 5 & Model 6 & Model 7 & Model 8\\
\hline
College Educated &  & 1.39$^{***}$ &  & 1.31$^{***}$ &  & 1.44$^{***}$ &  & 1.43$^{***}$\\
 &  & 0.02 &  & 0.01 &  & 0.02 &  & 0.02\\
 Median Income & 1.11$^{***}$ & 0.93$^{***}$ &  &  &  &  & 1.05$^{***}$ & 0.96$^{***}$\\
 & 0.01 & 0.01 &  &  &  &  & 0.02 & 0.01\\
Rural & 0.86$^{***}$ & 0.98 &  &  &  &  & 0.93$^{***}$ & 1\\
 & 0.01 & 0.01 &  &  &  &  & 0.01 & 0.01\\
 Unemployed & 1 & 1.01 &  &  &  &  & 0.98 & 1.01\\
 & 0.01 & 0.01 &  &  &  &  & 0.01 & 0.01\\
Black &  &  & 0.99 & 1 &  &  & 0.99 & 0.99\\
 &  &  & 0.01 & 0.01 &  &  & 0.01 & 0.01\\
 Hispanic &  &  & 0.99 & 1.04$^{***}$ &  &  & 0.98$^{***}$ & 1.04$^{***}$\\
 &  &  & 0.01 & 0.01 &  &  & 0.01 & 0.01\\
 Asian &  &  & 1.28$^{***}$ & 1.06$^{***}$ &  &  & 1.16$^{***}$ & 1.08$^{***}$\\
 &  &  & 0.02 & 0.01 &  &  & 0.02 & 0.01\\
Tech Share &  &  &  &  & 1.16$^{***}$ & 0.94$^{***}$ & 1.03 & 0.92$^{***}$\\
 &  &  &  &  & 0.01 & 0.01 & 0.02 & 0.01\\
Arts Share &  &  &  &  & 1.14$^{***}$ & 1 & 1.1$^{***}$ & 1\\
 &  &  &  &  & 0.01 & 0.01 & 0.01 & 0.01\\
Finance Share &  &  &  &  & 1.01 & 0.96$^{***}$ & 0.99 & 0.97$^{***}$\\
 &  &  &  &  & 0.01 & 0.01 & 0.01 & 0.01\\
Service Share &  &  &  &  & 0.96$^{***}$ & 1.03$^{***}$ & 1 & 1.02\\
 &  &  &  &  & 0.01 & 0.01 & 0.01 & 0.01\\
\hline
AIC                    & $35782$    & $35143$    & $35735$    & $35149$    & $35756$    & $35113$    & $35535$    & $35046$    \\
BIC                    & $35817$    & $35183$    & $35769$    & $35190$    & $35797$    & $35159$    & $35610$    & $35127$    \\
Log Likelihood         & $-17885$   & $-17564$   & $-17861$   & $-17567$   & $-17871$   & $-17548$   & $-17754$   & $-17509$   \\
Var: STATE (Intercept) & $0.02$        & $0.02$        & $0.02$        & $0.02$        & $0.02$        & $0.02$        & $0.02$        & $0.02$        \\
\hline
\end{tabular}
\caption{Results are exponentiated coefficients and standard errors from multilevel negative binomial models. Models predict the count of searches for ChatGPT, with the total count of searches by county included as an offset. N = 2,397 counties across 50 states and the District of Columbia.$^{***}p<0.001$; $^{**}p<0.01$; $^{*}p<0.05$ }
\label{table:coefficients}
\end{table*}

We next examine the associations between rates of search for chatGPT and socioeconomic, demographic, and industry makeup respectively. In Figure~\ref{fig4:all}, we see a strong positive association between the fraction college-educated or median incomes and logged rates of search. In particular, in counties in the top 10\% with respect to the fraction of residents who are college-educated, population-weighted median rates of search are approximately 7.7 per 10k searches (IQR 6.8 - 10.4) versus 2.8 (IQR 2.4-3.3) for the bottom 10\% of counties. We also observe strong positive associations with median income and fraction Asian. Notably, Figure~\ref{fig4:all} offers evidence of an inverse association with the fraction of the population that is non-Hispanic White, with counties with the highest rates of search characterized by proportionally larger Hispanic or Latino and Asian populations; the gradient with respect to fraction Black is largely flat, albeit with some evidence of lower rates of search at the top percentiles. These associations should be interpreted with caution, however, as they are likely confounded.

There is a strong and increasing association between the percentile of fraction tech sector jobs in a given county and its rate of search for ChatGPT. We also observe a positive association with respect to finance sector jobs, though the gradient is not as steep. Counties in the lowest ten percentiles of arts sector jobs have lower rates of search (3.0, IQR 2.5 - 3.7) in comparison to other counties (6.3, 4.6 - 7.8); the trend flattens across the upper percentiles. No similar gradient is observed for service sector jobs.

We next estimate associations in fully adjusted models. Before interpreting the coefficients, we assess model fit. Likelihood ratio tests confirm a significant improvement in fit from the inclusion of a random effect, as well as from the use of negative binomial rather than simpler poisson models to account for overdispersion; we also observe large reductions in AIC and BIC values for the multilevel negative binomial specification in comparison with simpler approaches. Residual tests show that observations do deviate from the distribution expected given the model, but not severely; in particular, we see no significant evidence of over- or under-dispersion or of outliers. Finally, spatial autocorrelation in the residuals is attenuated relative to that observed for the raw values (Moran's I = 0.18), though still statistically significant (p $<$ 0.001). These results indicate that, although some correlation is still unaccounted for, state-level random effects and included covariates explain a meaningful portion of the data's spatial structure.

Table~\ref{table:coefficients} shows rate ratios from negative binomial models with state-level random effects. We first examine the adjusted assocations between socioeconomic factors and rates of search for ChatGPT (columns 1 and 2). We continue to see negative associations of fraction rural and rates of search for ChatGPT after adjusting for median income (Column 1); however, after including education as a covariate, the fraction rural is no longer statistically significant and median income is actually inversely associated with rates of search for ChatGPT (Column 2).

Examining adjusted associations with demographic makeup, Table~\ref{table:coefficients} column 3 shows a strong positive association with the fraction Asian. A county that has 1 standard deviation (SD) higher fraction Asian would have approximately 1.3 times the rate of search for ChatGPT of a comparable county in the same state with the same Black and Hispanic fractions (implying a reduction in fraction White, which is excluded). This association remains positive and significant, though it is attenuated, after the inclusion of education as a covariate (Column 4).\footnote{Notably, there is a significant and negative association with the fraction Black in univariate models as well as in robustness checks with Google Trends data (Appendix), but this association is not significant in the main specifications after the inclusion of state-level random effects. As a supplementary analysis, we fitted models with percent White (excluding other demographic variables.) We find that the monotonic negative relationship with percent White observed in Figure~\ref{fig4:all} is robust to the inclusion of eduction but is no longer significant, in adjusted models, after accounting for urbanicity.}

Columns 5 and 6 offer evidence that education confounds the association of industry makeup and search rates. In column 5, we observe a significant and positive effect of share tech sector jobs, a smaller but still significant and positive association with arts sector jobs share, and a small but significant inverse association with the share of jobs in the service industry. However, after adjusting for fraction college educated, associations are attenuated or even inverted (Column 6). 

Finally, fully adjusted models follow a similar pattern (Columns 7 and 8). When education is excluded, we see positive associations with median income, fraction Asian, and arts share, and negative associations with fraction rural and fraction Hispanic. After accounting for education, we continue to see positive associations with fraction Asian and fraction Hispanic, but associations with median income, tech share, and finance share are negative and other associations are no longer significant. Notably, the association with education is both significant and large: holding other factors constant, a county with 1 SD higher fraction college educated would have approximately 1.4 times more searches for ChatGPT as a fraction of total searches. Our results are robust to the use of Google Trends rather than our proprietary search engine (See Appendix for details).

\section{Limitations}

Our study is subject to several limitations. First, we recognize that, as a proxy for actual usage, search rates may be subject to confounding over time if users use search only for initial discovery and then gradually select out of the data set. We mitigate this challenge by examining a completely novel tool (ChatGPT), and by limiting our analyses to only the first six months after it became publicly available---a period for which searches are the best available early data.

Second, we rely on data from a proprietary search engine in order to obtain a sample size large enough for fine-grained spatial analysis. Our findings could be subject to selection bias due to the set of people who choose to use the particular search provider from which we obtain our data. Moreover, to protect the privacy of search users, data are suppressed for some counties. However, results are similar when we conduct analyses with the publicly available Google Trends database, a search database with the highest rates of coverage nationally~\citep{statcounter2023}, but for which the finest available granularity is at the metropolitan-area level. 

Third, we examine spatially aggregated outcomes that may be subject to the modifiable areal unit problem in which aggregate patterns do not correspond directly with individual-level relationships~\citep{openshaw1984modifiable}. However, our findings are consistent with evidence from individual-level surveys~\citep{vogels2023chatgpt,ParkGellesWatnick2023, Motyl2024, McClain2024, bick2024rapid}. 

Finally, our findings are not causally identified and focus exclusively on a Global North context. To our knowledge, however, this study is the first national-scale characterization of geographic differences in uptake of generative AI, a prospective new general-purpose technology. The findings presented here support further research to better characterize and isolate the relevant processes of social and spatial diffusion that produce these aggregate patterns.  

\section{Discussion}\label{sec:discussion}

This study characterizes the digital divide with respect to knowledge of generative AI in the U.S. Examining county- and state-level rates of internet search for a popular new technology, ChatGPT, we make three contributions. First, we document emerging hot- and coldspots across the U.S. with respect to generative AI awareness. Second, we observe that counties with the highest rates of search for ChatGPT are relatively more urbanized, higher income, more educated, more Asian, and have more technology jobs relative to other counties. Finally, we show that educational attainment confounds individual-level associations with search rates, emerging as the strongest predictor of interest in fully adjusted models. These findings offer evidence of a new, emerging set of disparities in generative AI awareness that follows similar patterns to those of prior digital divides~\citep{van2006digital, hargittai2002second, scheerder2017determinants}. The stark differences we observe in generative AI awareness across places should be of particular concern to policymakers, given the strong economic benefits associated with early adoption of novel technologies~\citep{saxenian1996regional, autor2013growth, moretti2012new, moretti2022place}.

The spatial clustering we document aligns with usage patterns one might expect of tooling designed for and primarily used by highly educated, knowledge-economy workers~\citep{Shani2023}. Our findings are also consistent with two studies documenting similar emerging differences between advantaged versus disadvantaged countries~\citep{khowaja2024chatgpt, liu2024earth}. To our knowledge, however, this study is the first to empirically characterize awareness of generative AI across places within a country. 
By offering granular evidence of the regions and metropolitan areas where awareness lags most (versus least), our spatial clustering approach provides actionable insights to inform local economic development. This geographic specificity can help not only in understanding disparities, that is, but also in supporting policymakers to target places for intervention.

We document associations between the uptake of a novel technology that are expected (including positive associations with urbanicity, median income, and fraction technology jobs) and unexpected (including negative associations with fraction White). These associations are confounded by educational attainment and urbanicity and should be interpreted with caution, but they reflect noteworthy absolute differences in generative AI awareness across groups. We caution against interpreting these findings with a deficit framework, in which differences in awareness are attributed to the shortcomings of the affected groups rather than to systemic and structural factors~\citep{noble2018algorithms, stewart2024social}; indeed, the importance of educational attainment as a predictor of awareness points to the role of the systemic factors that also underlie education divides. Finally, our study focuses on the second-level divide in awareness and usage of a novel technology. Further research is needed to shed light on third-level divides in who benefits versus who is harmed by this novel tooling. We note that search logs might offer insight on third-level divides, enabling researchers to study e.g. how people learn about beneficial versus potentially harmful use; such disaggregations were not possible in our study given that searches were still relatively rare and we needed to aggregate data to protect privacy. Future research should go beyond awareness to study actual use and its impacts either with search logs or, preferably, logs of actual generative AI interactions~\citep{suri2024use, tamkin2024clio}.

This analysis shows that early awareness of a major new technology followed a similar path to that observed for prior digital divides, in which differences in the uptake of new technologies have tended to reinforce rather than remediate socioeconomic disparities~\citep{vandijk2005deepening, vanDeursen2019}. That generative AI would follow these patterns is not necessarily an inherent feature of the technology: individual-level and experimental research show the potential of generative AI tooling to reduce performance and productivity gaps between novices and experts~\citep{noy2023experimental, brynjolfsson2023generative, dell2023navigating}. Researchers have also noted generative AI's potential to support the ``leapfrogging'' by disadvantaged places into positions of relative advantage: multilingual capabilities make generative AI tools more widely accessible to non-English speakers in comparison with existing digital tooling~\citep{ahuja2023mega} and  the multi-modality of advanced models could support usage in places with low levels of traditional literacy~\citep{henneborn2023}. But active effort is required on the part of the technology's designers to mitigate differences in cost and quality between high- and low-resource languages, to develop accessible interfaces, and to ensure that the technology meets the needs of prospective users beyond knowledge workers; as well as on the part of advocates and policy makers, to raise awareness of the tooling, to educate people with respect to its effective and beneficial use, and to foster inclusive development that is tailored to the needs and use cases of marginalized groups.

Decisions about technology design and policy will thus be important in determining, over the longer run, whether generative AI reinforces versus mitigates prior divides. Improving access to computers and broadband has required mass campaigns, including free resources at public libraries and active interventions across schools~\citep{norris2001digital}; a similar effort may be called for in the context of the new generative AI divide.

\section{Ethical Statement}
This study was reviewed and approved by the Microsoft Research Institutional Review Board (protocol ID 10590). The primary data for this study are proprietary data from the Bing search engine, and the usage of de-identified and aggregated data for research purposes is included in the terms of use. To protect user privacy, we used only de-identified data that we aggregated geographically; we further suppressed any geographies with fewer than fifty users represented in the data. We also use data from the American Community Survey, obtained via the National Historical Geographic Information System~\citep{manson2023ipums}, which similarly relies on aggregation and anonymization to protect privacy. Our work complies with the data usage guidelines set forth by both the proprietary search provider and the U.S. Census Bureau, and no explicit consent was obtained as part of this study.

\section{Acknowledgements}

The authors are grateful for comments and suggestions from Jina Suh, Jake Hofman, Dan Goldstein, Sonia Jaffe, Daricia Wilkinson, and Sid Suri. The work was also improved by careful feedback from Scott Page and Seth Spielman. 

\bibliography{main}

\section{Paper Checklist}

\begin{enumerate}

\item For most authors...
\begin{enumerate}
    \item  Would answering this research question advance science without violating social contracts, such as violating privacy norms, perpetuating unfair profiling, exacerbating the socio-economic divide, or implying disrespect to societies or cultures?
    \answerYes{Yes}
  \item Do your main claims in the abstract and introduction accurately reflect the paper's contributions and scope?
    \answerYes{Yes}
   \item Do you clarify how the proposed methodological approach is appropriate for the claims made? 
    \answerYes{Yes}
   \item Do you clarify what are possible artifacts in the data used, given population-specific distributions?
    \answerYes{Yes, see limitations.}
  \item Did you describe the limitations of your work?
    \answerYes{Yes, see limitations.}
  \item Did you discuss any potential negative societal impacts of your work?
    \answerYes{Yes, we discuss and caution against reinforcing deficit frameworks in the discussion section..} 
      \item Did you discuss any potential misuse of your work?
    \answerNo{No, this is an empirical analysis and has negligible potential for misuse.} 
    \item Did you describe steps taken to prevent or mitigate potential negative outcomes of the research, such as data and model documentation, data anonymization, responsible release, access control, and the reproducibility of findings?
    \answerYes{Yes, we discuss measures taken to protect privacy in the Data section.} 
  \item Have you read the ethics review guidelines and ensured that your paper conforms to them?
    \answerYes{Yes.}
\end{enumerate}

\item Additionally, if your study involves hypotheses testing...
\begin{enumerate}
  \item Did you clearly state the assumptions underlying all theoretical results?
    \answerYes{Yes, see methods and results.}
  \item Have you provided justifications for all theoretical results?
    \answerYes{Yes, see methods and results.}
  \item Did you discuss competing hypotheses or theories that might challenge or complement your theoretical results?
    \answerYes{Yes, see limitations and discussion.}
  \item Have you considered alternative mechanisms or explanations that might account for the same outcomes observed in your study?
    \answerYes{Yes, see limitations.}
  \item Did you address potential biases or limitations in your theoretical framework?
    \answerYes{Yes, see limitations.} 
  \item Have you related your theoretical results to the existing literature in social science?
    \answerYes{Yes, see introduction and related literature.}
  \item Did you discuss the implications of your theoretical results for policy, practice, or further research in the social science domain?
    \answerYes{Yes, see discussion.}
\end{enumerate}

\item Additionally, if you are including theoretical proofs...
\begin{enumerate}
  \item Did you state the full set of assumptions of all theoretical results?
    \answerNA{NA}
	\item Did you include complete proofs of all theoretical results?
    \answerNA{NA}
\end{enumerate}

\item Additionally, if you ran machine learning experiments...
\begin{enumerate}
  \item Did you include the code, data, and instructions needed to reproduce the main experimental results (either in the supplemental material or as a URL)?
    \answerNA{NA}
  \item Did you specify all the training details (e.g., data splits, hyperparameters, how they were chosen)?
    \answerNA{NA}
     \item Did you report error bars (e.g., with respect to the random seed after running experiments multiple times)?
    \answerNA{NA}
	\item Did you include the total amount of compute and the type of resources used (e.g., type of GPUs, internal cluster, or cloud provider)?
    \answerNA{NA}
     \item Do you justify how the proposed evaluation is sufficient and appropriate to the claims made? 
    \answerNA{NA}
     \item Do you discuss what is ``the cost`` of misclassification and fault (in)tolerance?
    \answerNA{NA}
  
\end{enumerate}

\item Additionally, if you are using existing assets (e.g., code, data, models) or curating/releasing new assets, \textbf{without compromising anonymity}...
\begin{enumerate}
  \item If your work uses existing assets, did you cite the creators?
    \answerYes{Yes, see Data.}
  \item Did you mention the license of the assets?
    \answerNo{No, the primary data are proprietary.}
  \item Did you include any new assets in the supplemental material or as a URL?
    \answerNo{No, because the search log data in this study are proprietary.}
  \item Did you discuss whether and how consent was obtained from people whose data you're using/curating?
    \answerYes{Yes, see ethics statement.}
  \item Did you discuss whether the data you are using/curating contains personally identifiable information or offensive content?
    \answerYes{Yes, the de-identification is discussed in the data section.}
\item If you are curating or releasing new datasets, did you discuss how you intend to make your datasets FAIR (see \citet{fair})?
\answerNA{NA}
\item If you are curating or releasing new datasets, did you create a Datasheet for the Dataset (see \citet{gebru2021datasheets})? 
\answerNA{NA}
\end{enumerate}

\item Additionally, if you used crowdsourcing or conducted research with human subjects, \textbf{without compromising anonymity}...
\begin{enumerate}
  \item Did you include the full text of instructions given to participants and screenshots?
    \answerNA{NA}
  \item Did you describe any potential participant risks, with mentions of Institutional Review Board (IRB) approvals?
    \answerNA{NA}
  \item Did you include the estimated hourly wage paid to participants and the total amount spent on participant compensation?
    \answerNA{NA}
   \item Did you discuss how data is stored, shared, and deidentified?
   \answerNA{NA}
\end{enumerate}

\end{enumerate}

\section{Appendix: Comparison with Google Trends}

\begin{figure*}[htbp]
\centering
\includegraphics[width=0.9\textwidth]{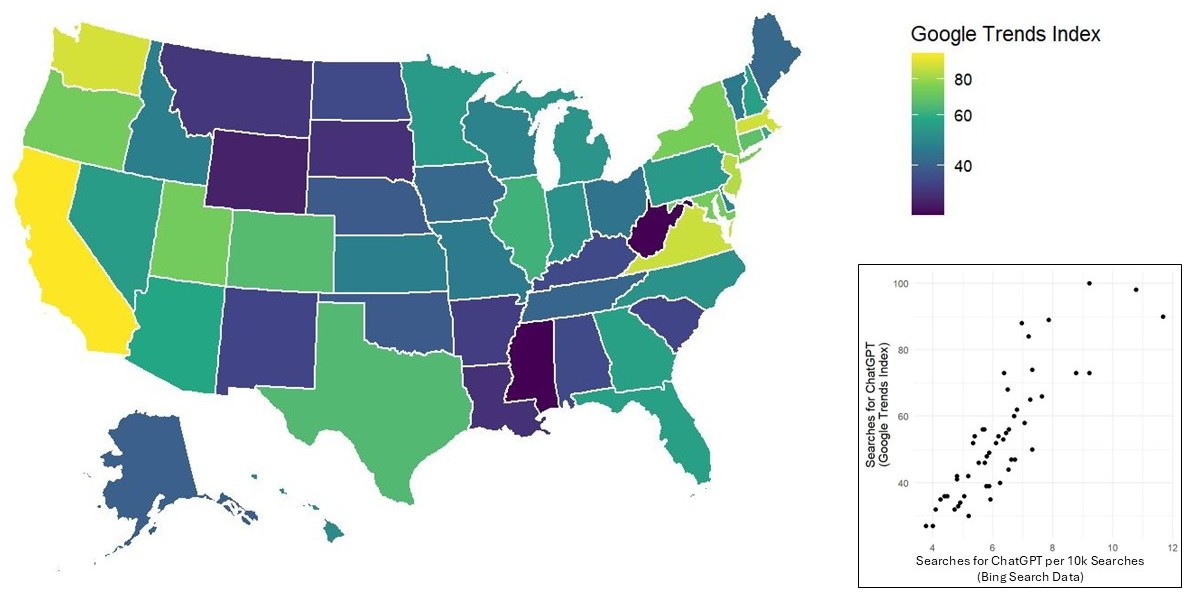}
\caption{Map of state-level searches for ChatGPT using the Google Trends Index. The inset shows that the Google Trends Index is highly correlated with the rates of search calculated from Bing search data (Pearson's Correlation = 0.86)}\label{figa1_google_map}
\end{figure*}

\begin{figure*}[htbp]%
\centering
\includegraphics[width=0.9\textwidth]{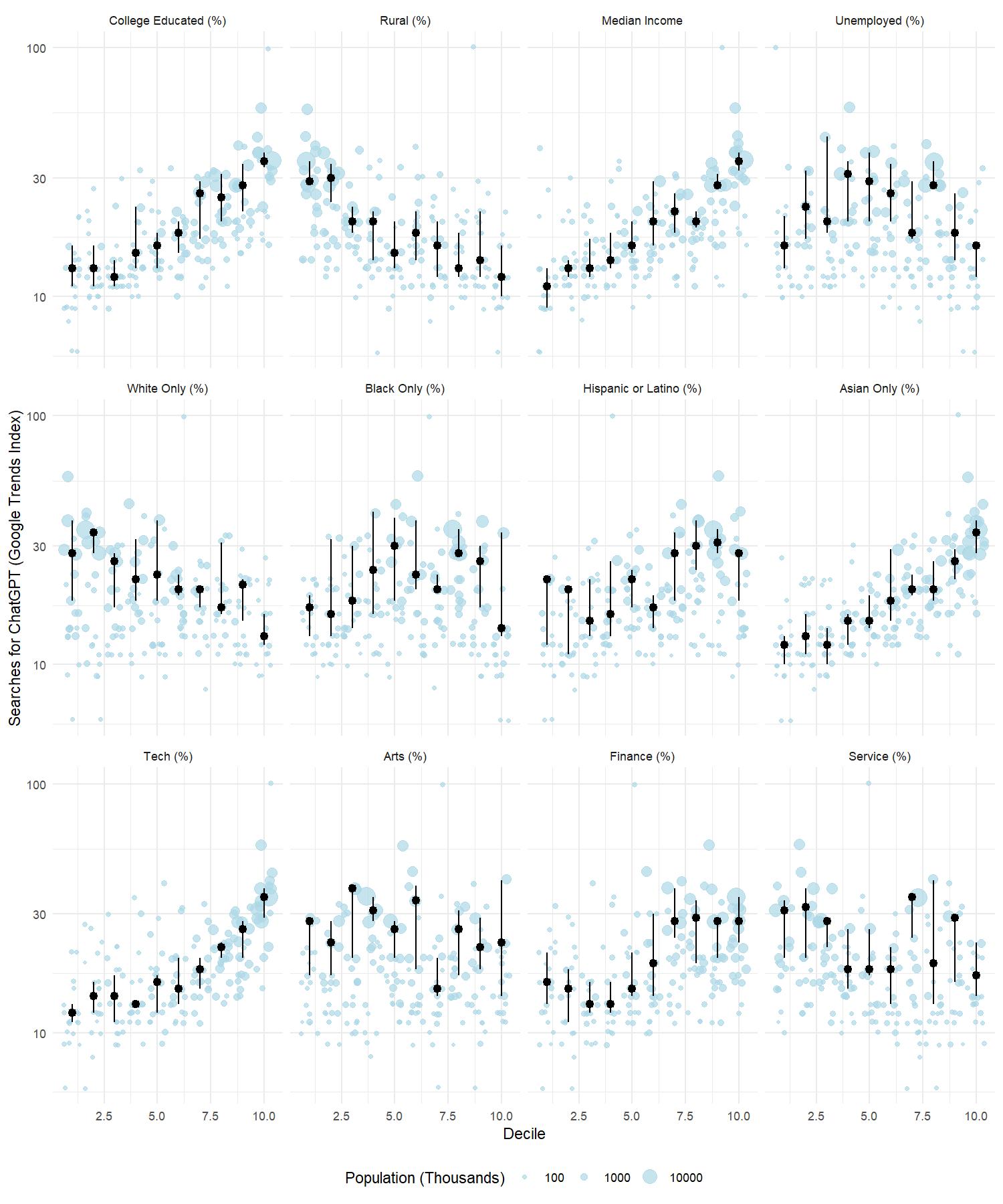}
\caption{Searches for ChatGPT in relation to socioeconomic, demographic, and industry makeup. Blue points indicate values from the Google Trends Index; black points are medians and 75\% interquartile ranges of metropolitan area ACS data, manually linked to the closest designated market area. Observations and estimates are weighted by metro area population.}\label{figa2:class_google}
\end{figure*}

Figure~\ref{figa1_google_map} compares state-level search trends in the Google Trends Index with estimates constructed using the Bing search data. Figure~\ref{figa2:class_google} shows univariate associations and Table~\ref{taba1:coefficients_google} reproduces the hierarchical models with the Google Trends Index.

\begin{table*}
\centering
\begin{tabular}[htbp]{lllllllll}
 & Model 1 & Model 2 & Model 3 & Model 4 & Model 5 & Model 6 & Model 7 & Model 8\\
\hline
(Intercept) & 17.52$^{***}$ & 17.1$^{***}$ & 17.74$^{***}$ & 17.26$^{***}$ & 17.59$^{***}$ & 17.18$^{***}$ & 17.58$^{***}$ & 17.28$^{***}$\\
(Intercept) & 0.62 & 0.6 & 0.63 & 0.53 & 0.65 & 0.59 & 0.53 & 0.52\\
College Educated &  & 1.34$^{***}$ &  & 1.28$^{***}$ &  & 1.32$^{***}$ &  & 1.33$^{***}$\\
                 &  & 0.05 &  & 0.03 &  & 0.05 &  & 0.05\\
Median Income   & 1.25$^{***}$ & 1.02 &  &  &  &  & 1.01 & 0.93\\
                & 0.04 & 0.04 &  &  &  &  & 0.05 & 0.04\\
Rural           & 0.95 & 0.98 &  &  &  &  & 1.03 & 1.01\\
                & 0.03 & 0.03 &  &  &  &  & 0.04 & 0.03\\
Unemployed      & 1 & 1.04 &  &  &  &  & 0.99 & 1.03\\
                & 0.03 & 0.03 &  &  &  & & 0.04 & 0.04\\
Black           &  &  & 0.93$^{***}$ & 0.95 &  &  & 0.92$^{***}$ & 0.92$^{***}$\\
                &  &  & 0.03 & 0.03 &  &  & 0.03 & 0.03\\
Hispanic        &  &  & 1.01 & 1.07$^{***}$ &  &  & 1.01 & 1.06\\
                &  &  & 0.03 & 0.03 &  &  & 0.04 & 0.04\\
Asian           &  &  & 1.28$^{***}$ & 1.1$^{***}$ &  &  & 1.14$^{***}$ & 1.12$^{***}$\\
                &  &  & 0.03 & 0.03 &  &  & 0.04 & 0.03\\
Tech Share      &  &  &  &  & 1.3$^{***}$ & 1.08$^{***}$ & 1.22$^{***}$ & 1.04\\
                &  &  &  &  & 0.04 & 0.04 & 0.05 & 0.04\\
Arts Share      &  &  &  &  & 0.99 & 0.96 & 0.98 & 0.97\\
                &  &  &  &  & 0.03 & 0.03 & 0.03 & 0.03\\
Finance Share   &  &  &  &  & 0.98 & 0.97 & 0.99 & 0.98\\
                &  &  &  &  & 0.03 & 0.02 & 0.03 & 0.03\\
Service Share   &  &  &  &  & 0.97 & 1.06 & 1.01 & 1.01\\
                &  &  &  &  & 0.04 & 0.04 & 0.05 & 0.04\\            
\hline
AIC                    & $1273$    & $1213$    & $1264$    & $1193$    & $1258$    & $1211$    & $1242$    & $1200$    \\
BIC                    & $1293$    & $1236$    & $1284$    & $1216$    & $1281$    & $1237$    & $1285$    & $1246$    \\
Log Likelihood         & $-631$   & $-599$   & $-17861$   & $-17567$   & $-17871$   & $-17548$   & $-17754$   & $-17509$   \\
Var: STATE (Intercept) & $0.03$        & $0.03$        & $0.02$        & $0.02$        & $0.02$        & $0.02$        & $0.02$        & $0.02$        \\
\hline
\end{tabular}
\caption{Results are exponentiated coefficients and standard errors from multilevel negative binomial models. Models predict the searches for ChatGPT (Google Search Index). N = 198 metro areas across 48 states. $^{***}p<0.001$; $^{**}p<0.01$; $^{*}p<0.05$} 
\label{taba1:coefficients_google}
\end{table*}

\end{document}